\pdfoutput=1

\documentclass[11pt]{article}

\usepackage{acl}

\usepackage{times}
\usepackage{latexsym}
\usepackage{booktabs}
\usepackage{algorithm}
\usepackage{algorithmic}
\usepackage{amsmath}
\usepackage{multirow}
\usepackage{times}
\usepackage[switch]{lineno}
\usepackage{natbib}

\usepackage[T1]{fontenc}

\usepackage[utf8]{inputenc}

\usepackage{microtype}

\usepackage{inconsolata}

\usepackage{graphicx}
\usepackage{amssymb}

\definecolor{blueyuhan}{RGB}{0,0,255}

\title{More is not always better? Enhancing Many-Shot In-Context Learning with Differentiated and Reweighting Objectives}

\author{
\begin{tabular}{c}
Xiaoqing Zhang$^{1,2}$\thanks{This work was done during the internship at Moonshot AI.} \quad \quad 
Ang\  Lv$^{1}$ \quad \quad 
Yuhan\  Liu$^{1}$ \quad \quad 
Flood Sung$^{2}$ \quad \quad \\
Wei Liu$^{3}$ \quad \quad Jian Luan$^{3}$ \quad \quad Shuo Shang$^{4}$ \quad \quad
Xiuying Chen$^{5}$\footnotemark[2] \quad \quad \ \ Rui \ Yan$^{1,6,7}$\thanks{\ \ Corresponding authors.} 
\end{tabular}
\\ \vspace{.5mm}
    \small
    \begin{tabular}{c}
    $^1$Gaoling School of Artificial Intelligence, Renmin University of China \quad $^2$MoonshotAI \quad
    $^3$Xiaomi AI Lab\\
    $^4$University of Electronic Science and Technology of China \quad
    $^5$Mohamed bin Zayed University of Artificial Intelligence\\
    $^6$School of Artifcial Intelligence, Wuhan University\\
    $^7$Engineering Research Center of Next-Generation Intelligent Search and Recommendation, MoE\\
    \end{tabular}
    \\ \vspace{.5mm}
    \small
    \begin{tabular}{c}
    \texttt{\{xiaoqingz, anglv, yuhan.liu, ruiyan\}@ruc.edu.cn} \quad \texttt{floodsung@moonshot.cn} \\ \texttt{\{liuwei40,luanjian\}@xiaomi.com} \quad \texttt{jedi.shang@gmail.com} \quad \texttt{xy-chen@pku.edu.cn}\\
    \end{tabular}
    \vspace{2mm} \\
}

\begin{document}
\maketitle
\begin{abstract}
Large language models (LLMs) excel at few-shot in-context learning (ICL) without requiring parameter updates. 
However, as ICL demonstrations increase from a few to many, performance tends to plateau and eventually decline. 
We identify two primary causes for this trend: the suboptimal negative log-likelihood (NLL) optimization objective and the incremental data noise. 
To address these issues, we introduce \textit{DrICL}, a novel optimization method that enhances model performance through \textit{Differentiated} and \textit{Reweighting} objectives. 
Globally, DrICL utilizes differentiated learning to optimize the NLL objective, ensuring that many-shot performance surpasses zero-shot levels. 
Locally, it dynamically adjusts the weighting of many-shot demonstrations by leveraging cumulative advantages inspired by reinforcement learning, thereby mitigating the impact of noisy data.
Recognizing the lack of multi-task datasets with diverse many-shot distributions, we develop the \textit{Many-Shot ICL Benchmark} (ICL-50)-a large-scale benchmark of 50 tasks that cover shot numbers from 1 to 350 within sequences of up to 8,000 tokens-for both fine-tuning and evaluation purposes.
Experimental results demonstrate that LLMs enhanced with DrICL achieve significant improvements in many-shot setups across various tasks, including both in-domain and out-of-domain scenarios.
We release the code and dataset hoping to facilitate further research in many-shot ICL\footnote{https://github.com/xiaoqzhwhu/DrICL}.
\end{abstract}

\begin{figure}[t]
    \centering
    \includegraphics[width=0.5\textwidth]{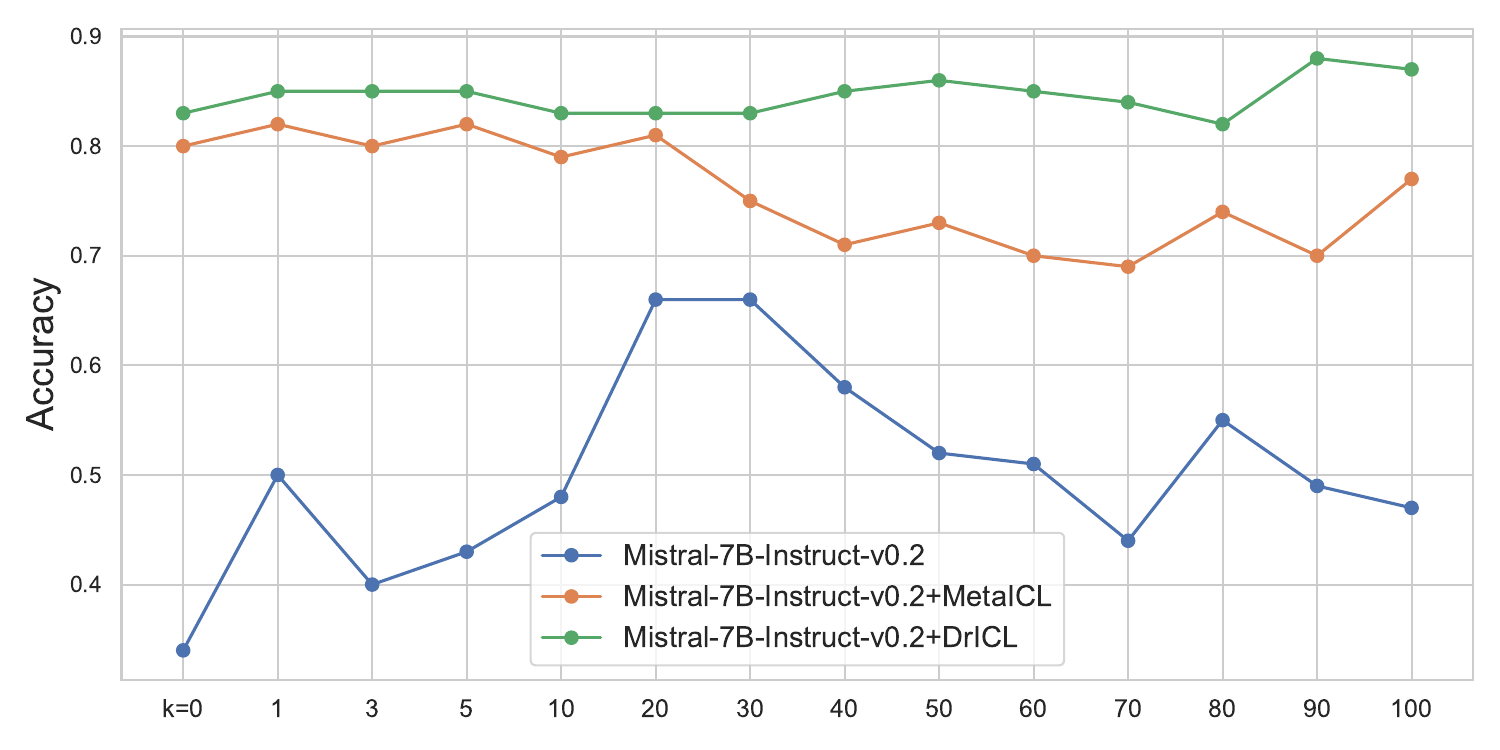}
    \caption{The performance trend of LLMs across different $k$-shots scenarios. $k$ refers to the number of demonstration examples provided to LLMs, ``+MetaICL'' uses MetaICL for fine-tuning, while ``+DrICL'' uses our DrICL strategy.}
    \label{fig:intro}
\end{figure}

\section{Introduction}
In-context learning~\cite{brown2020language} enables models to quickly adapt and address specific issues by utilizing contextual cues, improving adaptability and generalization.
With the expansion of the context length in advanced LLMs, the ability to process text lengths up to 1 million tokens allows LLMs to accept increasingly more demonstrations.
The ICL scenarios with hundreds or thousands of shots are called many-shot learning~\cite{agarwal2024many}.
However, many-shot does not always result in better performance than few-shot. 
Some models exhibit a linear decrease in ICL capabilities with the increase in ultra-long text lengths \cite{liu2024lost}. 
As shown in Figure \ref{fig:intro}, we present the accuracy variations of Mistral-7B-Instruct-v0.2 on the CLSClusteringS2S dataset \cite{li2022csl}. 
As the number of ICL examples increases, models' performance exhibits a trend of rising and then falling. 

We summarize two possible factors based on our preliminary study and previous works.
The first factor is the training objective. 
As ~\citeauthor{agarwal2024many} highlights, while the straightforward NLL decreases during testing with ICL, performance on many downstream tasks also deteriorates. 
The second factor is the increasing noise with the large number of demonstrations.
~\citeauthor{long2024decomposing, gao2024noise} demonstrate that the effectiveness of ICL heavily depends on the quality of the demonstrations.
While in many-shot scenarios, utilizing a large number of high-quality demonstrations presents significant challenges, such as the huge workload of creating them and the difficulty of domain adaptation.
Existing few-shot ICL methods do not address these issues, making them unsuitable for many-shot scenarios~\cite{zhang2024batch,li2023context,agarwal2024many,bertsch2024context}.

To address the above factors, we propose DrICL, enhancing many-shot in-context learning with a refined fine-tuning objective.
For the first factor, we propose differentiated learning to deal with the trade-off between many-shot and zero-shot scenarios from a \textit{global} perspective. 
During differentiated learning, we ensure that the performance on many-shot demonstrations surpasses that on zero-shot demonstrations. 
This approach promotes the model's deeper understanding of contextual cues, encouraging it to leverage contextual information effectively.
For the second factor, we find the noise in the demonstrations is reflected in the sharp increase in loss observed for certain samples, which disrupts the training process.
Inspired by reinforcement learning, we propose an advantage-based reweighting method to reduce focus on noise in many-shot demonstrations from a \textit{local} perspective.
In reinforcement learning, the ``advantage function'' is essential for assessing the value of actions beyond the average expected return, effectively directing the policy to choose actions that are predicted to yield the highest future rewards \cite{baird1994reinforcement}.
Similarly, just as the advantage function helps identify actions that yield higher returns than the average, we extend this concept by using cumulative advantage to adjust the weights of demonstrations.
This adjustment ensures that demonstrations with extreme loss fluctuations do not disproportionately influence the model or disrupt stable learning. 
The cumulative advantage for each demonstration is calculated based on the loss of the current demonstration and the losses of sampled demonstrations from preceding ones.
We introduce a sampling window to ensure a balanced consideration of previous demonstrations.
Each sequence is divided into multiple reweighting windows, and for each demonstration in a reweighting window $w$, the preceding window $w-1$ serves as the sampling window. 
Finally, we incorporate the cumulative advantage into the negative log-likelihood (NLL) computation, resulting in an advantage-based training objective.

In addition to the two challenges mentioned above, another significant obstacle is the lack of sufficient multi-task training data that spans a wide range of task numbers. 
Such data is crucial for studying the effects of ICL across many-shot scenarios. 
We present ICL-50, the largest dataset to study many-shot ICL, encompassing 50 tasks across 7 task types, and a total of over three million samples. 
The maximum number of shots achievable varies depending on the model, allowing for comprehensive investigations into many-shot ICL, including research on fine-tuning and inference.
We categorize ICL-50 into different subsets based on task types, including in-domain and out-of-domain tasks, to fully assess the model's ICL capabilities in many-shot scenarios.

In summary, this paper identified two challenges in avoiding ICL's decreasing performance under Many-shot scenarios: suboptimal training objective, and incremental data noise.
We propose ``differentiated learning'' and ``advantaged-based reweighting'' to address these challenges, respectively.
We further propose the largest ICL-50 to support our training, evaluation as well as further studies.
We experiment DrICL on the ICL-50 with open-source LLMs, showing stable performance both in-domain and out-of-domain under many-shots.
All these points form the major contribution of this paper.

\section{Related Work}
\textbf{In-context Learning.} In-context learning allows models to execute downstream tasks without the need for parameter updates, enabling language models to serve as a universal tool for a variety of tasks. 
As the number of examples supplied to LLMs grows, supplementary strategies become essential to bolster the model's ICL capabilities. 
For instance, \citet{anil2024many} employ multi-example prompts, which can accommodate up to 256 demonstrations, to overcome the inherent limitations of language models. \citet{hao2022structured} propose the structured prompting method to overcome length restrictions and extend in-context learning to thousands of examples. \citet{li2023context} use a customized model architecture to support the expansion of contextual examples to 2,000, and \cite{agarwal2024many} utilize reinforced ICL and unsupervised ICL to extend the scope of contextual examples to 8,192. 
Unlike their work, we enhance the ICL capability of LLMs by improving the model's parameters rather than the form of contextual examples. 

\textbf{Instruction Tuning of LLMs.} Instruction tuning has become an effective technique for enhancing the capabilities and controllability of LLMs \cite{zhang2023instruction}. 
In the domain of ICL, studies like MetaICL \cite{min2022metaicl} , IAD~\cite{liu2024iad}, and PEFT \cite{bertsch2024context} have demonstrated that fine-tuning LLMs with both small and large demonstration sizes, denoted as $k$, lead to improved ICL performance.
Despite their studies being confined to a modest quantity of tuning data—capped at 10,000 entries—there is an evident necessity for deeper research into how ICL performs when scaled up with more extensive datasets. 
Consequently, we introduce ICL-50, a significantly larger dataset, designed to delve into the strategies for amplifying ICL's potential.

\textbf{LLM Data Reweighting.} As LLMs rapidly advance, the application of data reweighting in training has become increasingly prevalent.
In the pre-training stage, SoftDedup significantly improves training efficiency by selectively reducing the sampling weight of data with high commonness through a soft deduplication method, rather than removing them to increase the integrity of the dataset \cite{he2024softdedup}. 
ScaleBiO reweights the data of LLMs by filtering irrelevant data samples and selecting informative samples, demonstrating its effectiveness and scalability across models of different sizes on tasks such as data denoising, multilingual training, and instruction tuning \cite{pan2024scalebio}. 
In the ICL scenario, \citet{yang2023not} propose WICL to enhance the performance of ICL by assigning optimal weights to demonstration samples in the inference. 
Unlike other works, we set the weights during the training process based on the positions of multiple examples in a sequence.

\begin{figure*}[htb]
    \centering
    \includegraphics[width=1.0\textwidth]{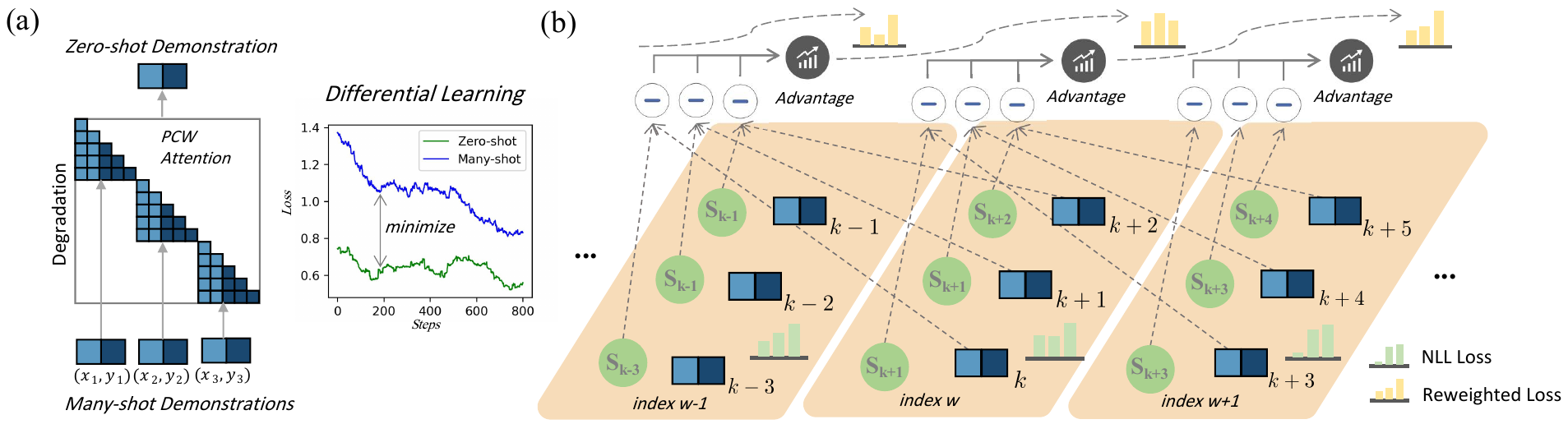} 
    \caption{The DrICL Training Framework. (a) The global differentiated learning for many-shot and zero-shot demonstrations. (b) The local advantage-based reweighting method assigns differential weights to demonstrations in window $w$ with window size $|W|=3$ and sampling size $|S|=1$, utilizing the cumulative advantage from the preceding window $w-1$.}
    \label{fig:icl}
\end{figure*}

\section{DrICL}
In this work, we propose the DrICL learning framework, which adjusts the weights of demonstrations and integrates reweighting within differentiated objectives, as illustrated in Figure~\ref{fig:icl}.
In DrICL, we organize training data through many-shot and zero-shot demonstrations. 
By simultaneously training the sequence of many-shot and zero-shot with a differentiated objective, we strengthen the model's overall ICL capability. 
At the same time, to further reduce the noise of demonstrations in many-shot scenarios, we introduce a weighted training objective towards different samples in the many-shot demonstrations. 
By sampling the model's performance under different demonstrations, we calculate the cumulative advantage gained as the number of demonstrations increases and use this cumulative advantage to adjust the learning process. 
Below, we show the components of the DrICL framework from both a global and local perspective, as well as the learning strategy.

\subsection{Global Perspective: Differentiated Learning}
We apply differentiated learning for the trade-off of many-shot and zero-shot sequences due to differing sample lengths, where longer sequences might introduce more noise.
We expect that after refining the learning objectives, the model can still perform well in scenarios with numerous demonstrations, longer samples, and potentially noisy backgrounds.
In each iteration, we sample $K$ pairs of examples $(x_k, y_k)$ from the training dataset, where $k$ ranges from 1 to $K$. Then, we concatenate the examples $x_k$ and their corresponding labels $y_k$, and the instruction $I$ generated by GPT-3.5-turbo for the current task as the input sequence $S_K = \{I; x_1y_1x_2y_2 \ldots x_Ky_K\}$. 
We train the model to predict the label $y_{k}$ of the $k$-th example based on the instruction and the features and labels of the previous $k-1$ examples. The training objective of the model is to minimize the NLL loss $\mathcal{L}_{\text{NLL}}$, with the previous $k-1$ examples as the training examples and the $k$-th example as the test example. This training method helps the model learn in context during the inference stage. 
We organize the number of demonstration examples according to $k$. When $k > 0$, we perform many-shot instruction-tuning, and when $k = 0$, we perform zero-shot instruction-tuning. 
During our training process, we expect that different examples of the same training sequence $S_k$ can serve as helpful contexts $C_h$ for each other. 
In the absence of context, we define $C_{\text{none}}$, so we update the original NLL loss combined with the additional objective for many-shot and zero-shot as follows:
\begin{align*}
\mathcal{L}_{\text{many-shot}} &= \mathcal{L}_{\text{NLL}}(\text{LLM}(C_h, Q; \theta), A_{gt}),\\
\mathcal{L}_{\text{zero-shot}} &= \mathcal{L}_{\text{NLL}}(\text{LLM}(C_{\text{none}}, Q; \theta), A_{gt}),
\end{align*}
where $Q$ is the input question to the model, and $A_{gt}$ is the corresponding ground truth answer. 
We utilized many-shot data as $Q$ and transformed it into a degraded zero-shot format using the Parallel Context Windows (PCW) method \cite{ratner2022parallel}. PCW works by masking many-shot sequences to generate a zero-shot sequence, effectively enabling us to leverage both formats. In our implementation, PCW was used solely to simplify the input for coding purposes.
We aim to simultaneously optimize these two losses, such that $\mathcal{L}_{\text{many-shot}} < \mathcal{L}_{\text{zero-shot}}$.
A lower many-shot loss signifies that the model has more effectively mastered in-context learning, thus enhancing its ability to accurately predict $A_{gt}$.

We have the following differentiated objectives:
\begin{align*}
\mathcal{L}_{\text{diff}} = (1 + \alpha) \ast \mathcal{L}_{\text{many-shot}} + (1 - \alpha) \ast \mathcal{L}_{\text{zero-shot}},
\end{align*}
where $\alpha$ is the hyperparameter that controls the trade-off between many-shot and zero-shot.

\subsection{Local Perspective: Advantage-based Reweighting}
After global Differential Learning, we noticed that loss fluctuates at certain $k$-shot points instead of decreasing consistently, suggesting some samples affect the model's context significantly, possibly introducing noise. 
To address this, we introduced a reweight mechanism that adjusts weights based on performance differences between adjacent windows, giving higher weights to samples with larger differences, and helping the model adapt to dynamic contexts.
The model optimizes the weights of demonstration data, continuously balancing exploration and data utilization to achieve a rapid and stable ICL performance.
Below, we describe the overall process from three aspects: importance sampling, advantage functions, and reweighting.

\subsubsection{Importance Sampling}
Importance sampling adjusts weights to reduce bias and imbalance from noisy data. Here, we use the training model's loss on samples to calculate their importance weights.
For each training sequence $S_K = \{x_1y_1x_2y_2 \ldots x_Ky_K\}$, we calculate the loss ${\mathcal{L}_{\text{many-shot}}}_k$ generated by the sequence $\{x_1y_1x_2y_2 \ldots x_k\}$ at the current $k$-th position to represent the features of $S_k$. 
Our goal is to weight examples by their significance, emphasizing critical instances and reducing focus on less important ones.

To prevent an undue focus on specific parts of the data, we introduce a reweighting window, designed to segment the sequence into multiple parts. 
Each window is intended to handle a portion of the sequence with a total length of $K$.
The sequence is segmented into $\lfloor \frac{K}{W} \rfloor$ equal windows, each with a size of $W$. 
For the $k$-th demonstration we have the reweighting window index $w$ as follows:
\begin{align*}
w = \left [ \lfloor \frac{k}{W} \rfloor \times W : \left( \lfloor \frac{k}{W} \rfloor + 1 \right)  \times W \right].
\end{align*}
We designate the preceding window $w-1$ as the sampling window, to select $|S|$ demonstrations for those in the reweighting window $w$, compiling these into a set $S$. The demonstrations within set $S$ are then utilized for further training, leveraging accumulated benefits to enhance learning.
\begin{align*}
w-1 = \left[ \left( \lfloor \frac{k}{W} \rfloor - 1 \right) \times W : \lfloor \frac{k}{W} \rfloor \times W \right].
\end{align*}

We define a target distribution $p(x)$ and an importance distribution $q(x)$ with their probability density functions to achieve set $S$.
Specifically, for each training sample $S_k$ and feature vector $\mathcal{L}_k$ of the $k$-th demonstrations, we use the ratio of the values of the target distribution $p(x)$ and the importance distribution $q(x)$ to calculate the weight $weight_{k}$ for the $k$-th demonstration in $S_k$ and select the top $|S|$ samples with the highest weights:
\begin{align*}
weight_{k} = \frac{p({\mathcal{L}_{\text{many-shot}}}_k)}{q({\mathcal{L}_{\text{many-shot}}}_k)}.
\end{align*}
Through these steps, we calculate the weights of important samples and select the top $|S|$ representative samples from the given sample distribution.

\begin{figure*}[htb]
    \centering
    \includegraphics[width=0.8\textwidth]{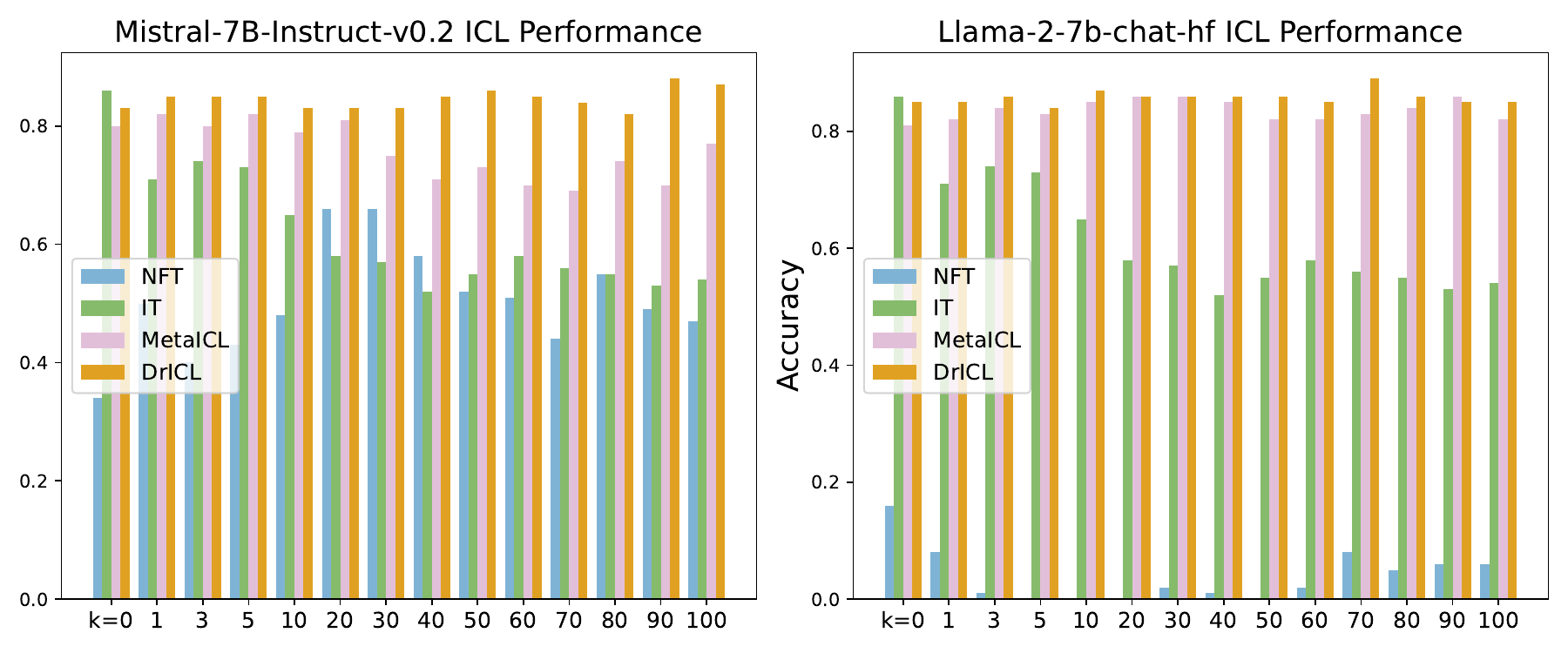} 
    \caption{The performance with incremental $k$-shots for Mistral-7B-Instruct-v0.2 and Llama-2-7b-chat-hf on CLSClusteringS2S under different strategies.
    We focus on CLSClusteringS2S for its high $k$-shot count, enabling a broader evaluation of DrICL.
    Our DrICL consistently shows better performance with a diverse range of $k$.}
    \label{fig:baselines}
\end{figure*}

\begin{table*}[htbp]
  \centering
  \small
  \resizebox{12cm}{!}{
\begin{tabular}{cccccccccccccc}
\toprule
\multicolumn{1}{c}{\multirow{2}[4]{*}{\textbf{Dataset}}} & \multirow{2}[4]{*}{\textbf{Models}} & \multicolumn{3}{c}{\textbf{k=0}} & \multicolumn{3}{c}{\textbf{k=1}} & \multicolumn{3}{c}{\textbf{k=3}} & \multicolumn{3}{c}{\textbf{k=5}} \\
\cmidrule{3-14}      & \multicolumn{1}{c}{} & \multicolumn{1}{c}{\textbf{D3}} & \multicolumn{1}{c}{\textbf{R1}} & \multicolumn{1}{c}{\textbf{B1}} & \multicolumn{1}{c}{\textbf{D3}} & \multicolumn{1}{c}{\textbf{R1}} & \multicolumn{1}{c}{\textbf{B1}} & \multicolumn{1}{c}{\textbf{D3}} & \multicolumn{1}{c}{\textbf{R1}} & \multicolumn{1}{c}{\textbf{B1}} & \multicolumn{1}{c}{\textbf{D3}} & \multicolumn{1}{c}{\textbf{R1}} & \multicolumn{1}{c}{\textbf{B1}} \\
\midrule
\multicolumn{1}{c}{\multirow{4}[2]{*}{XSUM$\stackrel{\text{id}}{}$}} & NFT   & 0.08  & 0.14  & 0.17  & 0.08  & 0.17  & 0.14  & 0.07  & 0.09  & 0.11  & 0.04  & 0.12  & 0.07 \\
      & IT    & 0.19  & 0.22  & 0.31  & 0.18  & 0.18  & 0.29  & 0.17  & 0.18  & 0.27  & 0.16  & 0.18  & 0.28 \\
      & MetaICL & 0.19  & 0.23  & 0.30   & 0.15  & 0.22  & 0.31  & 0.18  & 0.23  & 0.30   & 0.18  & 0.22  & 0.29 \\
      & DrICL & \textbf{0.20} & 0.22  & \textbf{0.33} & \textbf{0.19} & 0.21 & \textbf{0.32} & \textbf{0.20} & \textbf{0.23} & \textbf{0.34} & \textbf{0.20} & \textbf{0.22} & \textbf{0.33} \\
\midrule
\multicolumn{1}{c}{\multirow{4}[2]{*}{CNN$\stackrel{\text{ood}}{}$}} & NFT   & 0.07  & 0.29  & 0.20   & 0.06  & 0.21  & 0.18  & 0.04  & 0.15  & 0.09  & 0.05  & 0.07  & 0.05 \\
      & IT    & 0.18  & 0.34  & 0.51  & 0.18  & 0.32  & 0.51  & 0.15  & 0.27  & 0.39  & 0.20   & 0.11  & 0.14 \\
      & MetaICL & 0.19  & 0.34  & 0.51  & 0.19  & 0.35  & 0.51  & 0.18  & 0.34  & 0.51  & 0.18  & 0.33  & 0.47 \\
      & DrICL & \textbf{0.19} & \textbf{0.34} & \textbf{0.51} & \textbf{0.19} & 0.34  & \textbf{0.51} & \textbf{0.19} & 0.31  & \textbf{0.52} & 0.19  & 0.31  & \textbf{0.47} \\
\bottomrule
\end{tabular}%
    }
  \caption{Summarization results on Llama-2-7b-chat-hf, where ``id'' denotes in-domain datasets and ``ood'' signifies out-of-domain datasets.
  \textbf{Bold} indicates that our model performs the best.}
  \label{tab:summ}%
\end{table*}%

\begin{table}[htbp]
  \centering
  \resizebox{\linewidth}{!}{
\begin{tabular}{cccccccc}
\toprule
\multicolumn{1}{c}{\textbf{Dataset}} & \textbf{Models} & \multicolumn{1}{c}{\textbf{k=0}} & \multicolumn{1}{c}{\textbf{k=1}} & \multicolumn{1}{c}{\textbf{k=3}} & \multicolumn{1}{c}{\textbf{k=5}} & \multicolumn{1}{c}{\textbf{AVG}} & \multicolumn{1}{c}{\textbf{MAX}} \\
\midrule
\multicolumn{1}{c}{\multirow{4}[2]{*}{EcomRetrieval$\stackrel{\text{id}}{}$}} & NFT   & 0.19  & 0.10   & 0.09  & 0.01  & 0.10   & 0.19 \\
      & IT    & 0.93  & 0.06  & 0.12  & 0.19  & 0.33  & 0.93 \\
      & MetaICL & 0.89  & 0.85  & 0.92  & 0.91  & 0.89  & 0.92 \\
      & DrICL  & \textbf{0.93} & \textbf{0.87} & \textbf{0.94} & \textbf{0.93} & \textbf{0.92} & \textbf{0.94} \\
\midrule
\multicolumn{1}{c}{\multirow{4}[2]{*}{VideoRetrieval$\stackrel{\text{ood}}{}$}} & NFT   & 0.32  & 0.39  & 0.14  & 0.04  & 0.22  & 0.39 \\
      & IT    & 1.00     & 0.18  & 0.21  & 0.33  & 0.43  & 1.00 \\
      & MetaICL & 0.89  & 0.96  & 1.00     & 1.00     & 0.96  & 1.00 \\
      & DrICL  & \textbf{1.00} & \textbf{1.00} & \textbf{1.00} & \textbf{1.00} & \textbf{1.00} & \textbf{1.00} \\
\bottomrule
\end{tabular}%
    }
    \caption{Retrieval performance on Llama-2-7b-chat-hf. \textbf{Bold} indicates that our model performs the best.}
  \label{tab:retrieval}%
\end{table}%

\subsubsection{Advantage Functions} 
To assess the model's cumulative advantages as the $k$-shots grow, we select the sample set $S$ for the $k$-th instance within the weighting window $w$. Subsequently, we determine the average loss of the sampling window $w-1$ using the formula:
\vspace{-1mm}
\begin{equation*}
\begin{aligned}
\mathcal{L}_{\text{sampling}_{w-1}} = \frac{1}{|S|} \sum_{\text{instance}_i \in S} \mathcal{L}_{\text{instance}_i}.
\end{aligned}
\vspace{-2mm}
\end{equation*}
The reward is defined as the difference between the loss of the instance at the current position $k$ and the average loss of the instances in window $w-1$:
\vspace{-1mm}
\begin{equation*}
\begin{aligned}
\mathcal{R}_k = {\mathcal{L}_{\text{many-shot}}}_k - \mathcal{L}_{\text{sampling}_{w-1}}
\end{aligned}
\vspace{-2mm}
\end{equation*}
Here, $\mathcal{L}_{\text{sampling}_{w-1}}$ represents the performance of the model on all sampled instances before window $w$. 
It denotes the model's performance with fewer than $k$ shots, whereas ${\mathcal{L}_{\text{many-shot}}}_k$ signifies the model's performance with $k$ shots.
Next, we define the accumulated advantages to measure the strategy's performance in different positions $k$:
\begin{equation*}
\begin{aligned}
\mathcal{A}_k = \exp(\mathcal{R}_k / \gamma),
\end{aligned}
\vspace{-1mm}
\end{equation*}
where $\gamma$ is a temperature parameter used to adjust the sensitivity of the rewards. The exponential increase in the advantages metric strengthens positive rewards while suppressing negative rewards, guiding the model to select strategies that bring significant performance improvement.

\subsubsection{Reweighting}
In the DrICL framework, we select important samples in the previous window and calculate the reward that measures the model's performance in different positions to update the NLL loss for many-shot scenarios.
We adjust the overall training objective of the many-shot sequence as follows:
\begin{equation*}
\begin{aligned}
\mathcal{L}_{\text{many-shot}}=\frac{1}{k} \sum\limits_k{ {\mathcal{L}_{\text{many-shot}}}_k*\mathcal{A}_k}.
\end{aligned}
\vspace{-2mm}
\end{equation*}
By introducing the reweighting mechanism, we can not only maintain the performance of the current demonstration but also further optimize the model through gradient descent, leading to improved long-term performance.

\subsubsection{Learning Strategy}
The detailed process of the DrICL is presented in Algorithm \ref{alg:DrICL}. 
It enables the model to build upon prior knowledge at each iteration, avoiding uniform learning, thereby achieving progressive performance enhancement over long-term training. 

\begin{algorithm}[tb]
\small
\caption{Differentiated and Reweighting In-Context Learning (DrICL)}
\label{alg:DrICL}
\textbf{Parameter}: $\alpha$, $\gamma$, $S$, $W$
\begin{algorithmic}[1] 
\STATE Initialize training data $D$, total number of iterations $T$, set current iteration $t=0$.
\FOR{$t$ in $T$}
\FOR{d=$x_1,y_1,x_2,y_2,...,x_K,y_K$ in $D$}
\STATE Let the zero-shot loss $\mathcal{L}_{\text{zero-shot}}=0$, many-shot loss $\mathcal{L}_{\text{many-shot}}=0$.
\FOR {$k$ in $K$}
\STATE Calculate the many-shot loss ${\mathcal{L}_{\text{many-shot}}}_k$.
\STATE Mask the context of $x_k$ by PCW attention to get the sequence $\text{zero-shot}_k$.
\STATE Calculate the zero-shot loss ${\mathcal{L}_{\text{zero-shot}}}_k$.
\STATE Set the window index $w=\lfloor{k}/{W}\rfloor$.
\STATE Sample $|S|$ demonstrations from the window $w-1$ based on importance to form a validation set $S$.
\STATE Calculate the sampling loss $\mathcal{L}_{\text{sampling}_{w-1}}$ for the demonstrations in $S$.
\STATE Set the $\mathcal{R}_k = {\mathcal{L}_{\text{many-shot}}}_k - \mathcal{L}_{\text{sampling}_{w-1}}$.
\STATE Update the cumulative advantage: $\mathcal{A}_k = exp\left(\mathcal{R}_k/\gamma\right)$.
\STATE Assign the weighted loss: ${\mathcal{L}_{\text{many-shot}}}_k  = {\mathcal{L}_{\text{many-shot}}}_k  \times \mathcal{A}_k $.
\STATE $\mathcal{L}_{\text{many-shot}}\mathrel{+}={\mathcal{L}_{\text{many-shot}}}_k$.
\STATE $\mathcal{L}_{\text{zero-shot}}\mathrel{+}={\mathcal{L}_{\text{zero-shot}}}_k$.
\ENDFOR
\STATE $\mathcal{L}_{\text{many-shot}}=\mathcal{L}_{\text{many-shot}}/{K}$.
\STATE $\mathcal{L}_{\text{zero-shot}}=\mathcal{L}_{\text{zero-shot}}/{K}$.
\STATE Update $\mathcal{L}_{\text{diff}}$ with hyperparameter $\alpha$.
\ENDFOR
\ENDFOR
\end{algorithmic}
\end{algorithm}

\begin{table*}[h]
  \small
  \centering
  \resizebox{16cm}{!}{
\begin{tabular}{ccccccccccccccc}
\toprule
\multicolumn{1}{c}{\textbf{Dataset}} & \textbf{Models} & \multicolumn{1}{c}{\textbf{k=0}} & \multicolumn{1}{c}{\textbf{k=1}} & \multicolumn{1}{c}{\textbf{k=3}} & \multicolumn{1}{c}{\textbf{k=5}} & \multicolumn{1}{c}{\textbf{k=10}} & \multicolumn{1}{c}{\textbf{k=20}} & \multicolumn{1}{c}{\textbf{k=30}} & \multicolumn{1}{c}{\textbf{k=40}} & \multicolumn{1}{c}{\textbf{k=50}} & \multicolumn{1}{c}{\textbf{k=60}} & \multicolumn{1}{c}{\textbf{k=70}} & \multicolumn{1}{c}{\textbf{AVG}} & \multicolumn{1}{c}{\textbf{MAX}} \\
\midrule
\multicolumn{1}{c}{\multirow{4}[2]{*}{OpenbookQA$\stackrel{\text{id}}{}$}} & NFT   & 0.27  & 0.28  & 0.30   & 0.28  & 0.26  & 0.22  & 0.21  & 0.23  & 0.19  & 0.21  & 0.13  & 0.23  & 0.28 \\
      & IT    & 0.70   & 0.50   & 0.57  & 0.56  & 0.57  & 0.59  & 0.56  & 0.54  & 0.54  & 0.50   & 0.44  & 0.55  & 0.70 \\
      & MetaICL & 0.59  & 0.59  & 0.64  & 0.63  & 0.72  & 0.75  & 0.77  & 0.78  & 0.78  & 0.79  & 0.70   & 0.70   & 0.79 \\
      & DrICL & 0.69  & \textbf{0.72} & \textbf{0.77} & \textbf{0.77} & \textbf{0.78} & \textbf{0.76} & 0.76  & 0.76  & \textbf{0.80} & 0.76  & \textbf{0.76} & \textbf{0.76} & \textbf{0.80} \\
\midrule
\multicolumn{1}{c}{\multirow{4}[2]{*}{ARC$\stackrel{\text{ood}}{}$}} & NFT   & 0.65  & 0.31  & 0.23  & 0.16  & 0.29  & 0.22  & 0.25  & 0.19  & 0.11  & 0.10   & 0.03  & 0.23  & 0.65 \\
      & IT    & 0.71  & 0.60   & 0.62  & 0.62  & 0.60   & 0.60   & 0.60   & 0.57  & 0.60   & 0.30   & 0.21  & 0.55  & 0.71 \\
      & MetaICL & 0.67  & 0.71  & 0.76  & 0.78  & 0.76  & 0.78  & 0.78  & 0.82  & 0.82  & 0.78  & 0.71  & 0.76  & 0.82 \\
      & DrICL & \textbf{0.78} & \textbf{0.78} & \textbf{0.76} & \textbf{0.81} & \textbf{0.80} & \textbf{0.80} & \textbf{0.81} & 0.78  & 0.79  & 0.75  & 0.67  & \textbf{0.78} & 0.81 \\
\midrule
\multicolumn{1}{c}{\multirow{4}[2]{*}{CLSClusteringS2S$\stackrel{\text{id}}{}$}} & NFT   & 0.16  & 0.08  & 0.01  & 0.00     & 0.00     & 0.00     & 0.02  & 0.01  & 0.00     & 0.02  & 0.08  & 0.03  & 0.16 \\
      & IT    & 0.86  & 0.71  & 0.74  & 0.73  & 0.65  & 0.58  & 0.57  & 0.52  & 0.55  & 0.58  & 0.56  & 0.64  & 0.86 \\
      & MetaICL & 0.81  & 0.82  & 0.84  & 0.83  & 0.85  & 0.86  & 0.86  & 0.85  & 0.82  & 0.82  & 0.83  & 0.84  & 0.86 \\
      & DrICL & 0.85  & \textbf{0.85} & \textbf{0.86} & \textbf{0.84} & \textbf{0.87} & \textbf{0.86} & \textbf{0.86} & \textbf{0.86} & \textbf{0.86} & \textbf{0.85} & \textbf{0.89} & \textbf{0.86} & \textbf{0.89} \\
\midrule
\multicolumn{1}{c}{\multirow{4}[2]{*}{ArxivClusteringS2S$\stackrel{\text{ood}}{}$}} & NFT   & 0.04  & 0.05  & 0.09  & 0.11  & 0.08  & 0.05  & 0.01  & 0.03  & 0.04  & 0.06  & 0.00     & 0.05  & 0.11 \\
      & IT    & 0.39  & 0.32  & 0.25  & 0.29  & 0.20   & 0.23  & 0.22  & 0.20   & 0.18  & 0.21  & 0.21  & 0.25  & 0.39 \\
      & MetaICL & 0.35  & 0.41  & 0.36  & 0.33  & 0.42  & 0.36  & 0.37  & 0.39  & 0.33  & 0.37  & 0.39  & 0.37  & 0.42 \\
      & DrICL & 0.34  & 0.35  & \textbf{0.38} & \textbf{0.41} & 0.36  & \textbf{0.40} & \textbf{0.43} & 0.36  & \textbf{0.34} & \textbf{0.41} & 0.35  & \textbf{0.38} & \textbf{0.43} \\
\midrule
\multicolumn{1}{c}{\multirow{4}[2]{*}{TenkgnadClusteringS2S$\stackrel{\text{ood}}{}$}} & NFT   & 0.29  & 0.00     & 0.00     & 0.00     & 0.00     & 0.00     & 0.00     & 0.00     & 0.00     & 0.00     & 0.00     & 0.03  & 0.29 \\
      & IT    & 0.29  & 0.20   & 0.16  & 0.19  & 0.19  & 0.12  & 0.15  & 0.13  & 0.18  & 0.17  & 0.15  & 0.18  & 0.29 \\
      & MetaICL & 0.24  & 0.19  & 0.23  & 0.21  & 0.24  & 0.23  & 0.23  & 0.26  & 0.27  & 0.27  & 0.25  & 0.24  & 0.27 \\
      & DrICL & 0.23  & \textbf{0.23} & \textbf{0.23} & \textbf{0.25} & \textbf{0.30} & \textbf{0.26} & \textbf{0.27} & 0.23  & \textbf{0.27} & 0.26  & \textbf{0.26} & \textbf{0.25} & \textbf{0.30} \\
\bottomrule
\end{tabular}%
    }
    \caption{The performance of question answering, clustering, and classification tasks across various datasets on the Llama-2-7b-chat-hf model. \textbf{Bold} indicates that our model performs the best.
}
  \label{tab:all}%
\end{table*}%

\begin{table*}[htbp]
  \centering
  \resizebox{16cm}{!}{
    \begin{tabular}{ccccccccccccccccccc}
    \toprule
    \multicolumn{1}{c}{\textbf{Dataset}} & \textbf{Models} & \multicolumn{1}{c}{\textbf{k=0}} & \multicolumn{1}{c}{\textbf{k=1}} & \multicolumn{1}{c}{\textbf{k=3}} & \multicolumn{1}{c}{\textbf{k=5}} & \multicolumn{1}{c}{\textbf{k=10}} & \multicolumn{1}{c}{\textbf{k=20}} & \multicolumn{1}{c}{\textbf{k=30}} & \multicolumn{1}{c}{\textbf{k=40}} & \multicolumn{1}{c}{\textbf{k=50}} & \multicolumn{1}{c}{\textbf{k=60}} & \multicolumn{1}{c}{\textbf{k=70}} & \multicolumn{1}{c}{\textbf{k=80}} & \multicolumn{1}{c}{\textbf{k=90}} & \textbf{k=100} & \textbf{k=200} & \textbf{AVG} & \textbf{MAX} \\
    \midrule
    \multicolumn{1}{c}{\multirow{4}[2]{*}{CLSClusteringS2S}} & NFT   & 0.34  & 0.50   & 0.40   & 0.43  & 0.48  & 0.66  & 0.66  & 0.58  & 0.52  & 0.51  & 0.44  & 0.55  & 0.49  & 0.47  & 0.00     & 0.47  & 0.66 \\
          & IT    & 0.86  & 0.71  & 0.74  & 0.73  & 0.65  & 0.58  & 0.57  & 0.52  & 0.55  & 0.58  & 0.56  & 0.55  & 0.53  & 0.54  & 0.07  & 0.58  & 0.86 \\
          & MetaICL & 0.80   & 0.82  & 0.80   & 0.82  & 0.79  & 0.81  & 0.75  & 0.71  & 0.73  & 0.70   & 0.69  & 0.74  & 0.70   & 0.77  & 0.73  & 0.76  & 0.82 \\
          & DrICL & 0.83  & \textbf{0.85} & \textbf{0.85} & \textbf{0.85} & \textbf{0.83} & \textbf{0.83} & \textbf{0.83} & \textbf{0.85} & \textbf{0.86} & \textbf{0.85} & \textbf{0.84} & \textbf{0.82} & \textbf{0.88} & \textbf{0.87} & 0.71  & \textbf{0.84} & \textbf{0.88} \\
    \midrule
    \multicolumn{1}{c}{\multirow{4}[2]{*}{TweetSentimentExtraction}} & NFT   & 0.43  & 0.30   & 0.31  & 0.36  & 0.33  & 0.35  & 0.34  & 0.40   & 0.27  & 0.27  & 0.20   & 0.35  & 0.39  & 0.38  & 0.30   & 0.33  & 0.43 \\
          & IT    & 0.74  & 0.56  & 0.67  & 0.52  & 0.66  & 0.65  & 0.69  & 0.56  & 0.64  & 0.65  & 0.70   & 0.68  & 0.70   & 0.68  & 0.69  & 0.65  & 0.74 \\
          & MetaICL & 0.75  & 0.77  & 0.80   & 0.77  & 0.78  & 0.79  & 0.78  & 0.81  & 0.73  & 0.75  & 0.79  & 0.76  & 0.73  & 0.70   & 0.51  & 0.75  & 0.81 \\
          & DrICL & \textbf{0.82} & \textbf{0.81} & \textbf{0.81} & \textbf{0.80} & \textbf{0.83} & \textbf{0.80} & \textbf{0.80} & 0.78  & \textbf{0.83} & \textbf{0.78} & \textbf{0.79} & \textbf{0.76} & \textbf{0.77} & \textbf{0.81} & \textbf{0.76} & \textbf{0.80} & \textbf{0.83} \\
    \bottomrule
    \end{tabular}}%
    \caption{The performance variation of datasets with the highest $k$-shots on Mistral-7B-Instruct-v0.2. \textbf{Bold} indicates that our model performs the best.}
   
  \label{tab:mistral}%
\end{table*}%

\section{Experiments}

\subsection{Experimental Setup}
\subsubsection{Datasets} 
To delve into the exploration of many-shot ICL in LLMs, we need plenty of data across a wide range of $k$-shots. 
The datasets employed in MetaICL, like CROSSFIT \cite{ye2021crossfit} and UNIFIEDQA \cite{khashabi2020unifiedqa}, have a notable constraint: their task lengths are generally centered around 100 tokens. This focus restricts the wide range of $k$-shot distributions, especially when the training sequence length is constrained.
On the other hand, the LongICLBench dataset introduced by \citet{li2024long} significantly extends the length ranging from 1,000 to 50,000 tokens. 
Nonetheless, the dataset's limitation to a few hundred task instances renders it more suitable for inference rather than extensive training.
In light of these limitations, we have developed the ICL-50 dataset. 
It encompasses 7 tasks of diverse difficulty levels and includes 50 datasets with average sample lengths per task that vary from 10 to 14,000 tokens. 
With the number of samples extending from the hundreds into the hundreds of thousands, the ICL-50 dataset ensures a substantial volume of data suitable for training and inference.
More details can be found in the Appendix.

\subsubsection{Base Models}
We perform our experiments using two foundational models, namely Llama-2-7b-chat-hf and Mistral-7B-Instruct-v0.2.
The base models are trained by different paradigms:
\noindent $\bullet$ \textbf{NFT}~\cite{touvron2023llama, jiang2023mistral}: The foundational models with No Fine-tuning.
\noindent $\bullet$ \textbf{IT} \cite{wei2021finetuned}: Instruction Tuning foundational models with zero-shot examples.
\noindent $\bullet$ \textbf{MetaICL} \cite{min2022metaicl}: Fine-tuning foundational models with many-shot examples.

\subsubsection{Evaluation Metrics} In our evaluation, we employ accuracy for assessing the performance of question answering, clustering, logical reasoning, classification, and retrieval tasks. For the summarization task, we utilize Distinct of trigram tokens (D3), ROUGE for unigrams (R1), and BLEU for unigrams (B1) as our metrics. In the case of reranking tasks, we apply standard ranking metrics, including Precision at k $P@k$, Recall at k $R@k$, and Normalized Discounted Cumulative Gain $G@k$.

\subsubsection{Implementation Details}
For the Llama-2-7b-chat-hf model, we configured the hyperparameter $\alpha$ to 0.2, while for Mistral-7B-Instruct-v0.2, we set $\alpha$ to 0.4. 
We set the parameter $\gamma$ to counteract the effects of weight explosion, and our experiments identified 11 as the optimal value for this parameter.
We determined that the optimal sampling size for $S$ is 1, with the reweighted window size $W$ set at 10.
For details on the experimental hyperparameter settings, please refer to Appendix \ref{exp:parameter}.
For all training and evaluation tasks, we utilized 8 A100 GPUs.

\begin{table}[htbp]
  \centering
  \resizebox{\linewidth}{!}{
\begin{tabular}{cccccccccc}
\toprule
\multicolumn{1}{c}{\textbf{Dataset}} & \textbf{Models} & \multicolumn{1}{c}{\textbf{k=0}} & \multicolumn{1}{c}{\textbf{k=1}} & \multicolumn{1}{c}{\textbf{k=3}} & \multicolumn{1}{c}{\textbf{k=5}} & \multicolumn{1}{c}{\textbf{k=10}} & \multicolumn{1}{c}{\textbf{k=20}} & \multicolumn{1}{c}{\textbf{AVG}} & \multicolumn{1}{c}{\textbf{MAX}} \\
\midrule
\multicolumn{1}{c}{\multirow{4}[2]{*}{GSM8K}} & NFT   & 0.28  & 0.16  & 0.11  & 0.07  & 0.01  & 0.02  & 0.11  & 0.28 \\
      & IT    & 0.31  & 0.26  & 0.26  & 0.21  & 0.26  & 0.16  & 0.24  & 0.31 \\
      & MetaICL & 0.24  & 0.26  & 0.26  & 0.24  & 0.24  & 0.26  & 0.25  & 0.26 \\
      & DrICL & 0.30   & \textbf{0.28} & \textbf{0.31} & \textbf{0.27} & \textbf{0.32} & \textbf{0.26} & \textbf{0.29} & \textbf{0.32} \\
\bottomrule
\end{tabular}%
    }
  \caption{The reasoning performance on GSM8K for the Llama-2-7b-chat-hf model. \textbf{Bold} indicates that our model performs the best.}
  \label{tab:reason}%
\end{table}%

\subsection{Results of Tasks}
We validate our method on 12 datasets with both in-domain and out-of-domain tasks. Figure \ref{fig:baselines} compares baseline models on the CLSClusteringS2S dataset across different $k$-shots of Llama-2-7b-chat-hf and Mistral-7B-Instruct-v0.2. Table \ref{tab:summ} shows summarization performance, while Table \ref{tab:retrieval} details retrieval metrics. Results for question answering, clustering, and classification are summarized in Table \ref{tab:all}, and Table \ref{tab:reason} presents reasoning task performance. Our reranking experiments are shown in Table \ref{tab:ranking}. Given Mistral-7B-Instruct-v0.2’s superior performance on sequences over 4,000 tokens, we compare baseline variations across $k$-shots for the tasks with the highest $k$, as detailed in Table \ref{tab:mistral}.

As shown in Tables \ref{tab:summ}, \ref{tab:retrieval}, \ref{tab:all}, \ref{tab:reason}, and \ref{tab:ranking}, DrICL significantly improves performance across various tasks. While MetaICL shows substantial fluctuations in $k$-shot performance on datasets like OpenbookQA and ARC, DrICL maintains more stable results. The slight advantage of our method over Meta-ICL is due to its focus on optimizing many-shot loss. IT's performance declines with increasing context length, as it relies solely on zero-shot, which is less effective in many-shot scenarios. 
Additionally, Llama-2-7b-chat-hf's 4,000-token limit causes performance on the ARC dataset to drop from 0.82 to 0.78 when $k$ exceeds 50.
Under the DrICL framework, among the 12 datasets tested, $k=0$ led to a performance decrease on 5 datasets, no change on 2, and improvement on 5.
Overall, performance improved by 0.5\% with $k=0$, remaining stable. For $k>0$, datasets like CLSClusteringS2S showed continuous improvement, while DrICL effectively maintained performance stability as $k$ increased across most datasets.

\begin{table}[htbp]
  \centering
    \resizebox{\linewidth}{!}{
    \begin{tabular}{ccccccc}
    \toprule
    \multirow{2}[4]{*}{\textbf{cMedQA}} & \multicolumn{3}{c}{\textbf{k=0}} & \multicolumn{3}{c}{\textbf{k=1}} \\
\cmidrule{2-7}    \multicolumn{1}{c}{} & \textbf{P@10} & \textbf{R@10} & \textbf{G@10} & \textbf{P@10} & \textbf{R@10} & \textbf{G@10} \\
    \midrule
    MetaICL & \multicolumn{1}{c}{0.33} & \multicolumn{1}{c}{0.51} & \multicolumn{1}{c}{0.53} & \multicolumn{1}{c}{0.30} & \multicolumn{1}{c}{0.46} & \multicolumn{1}{c}{0.52} \\
    DrICL & 0.31 & 0.48 & \textbf{0.55} & \textbf{0.33} & \textbf{0.51} & \textbf{0.54} \\
    \bottomrule
    \end{tabular}}%
    \caption{The comparison of ranking performance on the cMedQA dataset for the Llama-2-7b-chat-hf model, with a focus on zero-shot and one-shot settings due to its handling of examples with an extensive number of tokens. \textbf{Bold} indicates that our model performs the best.}
  \label{tab:ranking}%
\end{table}%

\subsection{In-Context Learning Analysis}
\subsubsection{Performance Tradeoff}
We observe that the foundation models underperform on both datasets. After fine-tuning, the IT strategy achieves its best in the few-shot. MetaICL, benefiting from many-shot training data, performs well at larger $k$-shot levels but still shows significant fluctuations. In contrast, DrICL delivers more stable results, with accuracy steadily improving as $k$ increases. DrICL not only outperforms MetaICL in many-shot scenarios but also demonstrates faster loss convergence, as shown in Figure \ref{fig:learning}(a) in the Appendix \ref{exp:parameter}, indicating its tradeoff of many-shot and zero-shot demonstrations.

\begin{table}[htbp]
  \centering
  \resizebox{\linewidth}{!}{
\begin{tabular}{ccccc}
\toprule
\textbf{Dataset} & \textbf{NFT} & \textbf{IT} & \textbf{MetaICL} & \textbf{DrICL} \\
\midrule
OpenbookQA & 2.20E-03 & 3.90E-03 & 5.60E-03 & 8.00E-04 \\
ARC   & 2.41E-02 & 2.10E-02 & 2.00E-03 & 1.40E-03 \\
CLSClusteringS2S & 2.40E-03 & 1.00E-02 & 3.00E-04 & 2.00E-04 \\
ArxivClusteringS2S & 1.00E-03 & 3.70E-03 & 8.00E-04 & 9.00E-04 \\
TengkgnadClusteringS2S & 7.00E-03 & 1.90E-03 & 5.00E-04 & 5.00E-04 \\
TweetSentimentExtraction & 3.30E-03 & 3.40E-03 & 4.90E-03 & 5.00E-04 \\
GSM8K & 8.50E-03 & 2.20E-03 & 1.00E-04 & 5.00E-04 \\
XSUM  & 1.40E-03 & 2.20E-03 & 5.00E-05 & 5.00E-05 \\
CNN   & 3.90E-03 & 2.30E-02 & 3.00E-03 & 3.70E-03 \\
EcomRetrieval & 4.10E-03 & 1.20E-01 & 7.00E-04 & 8.00E-04 \\
VideoRetrieval & 2.00E-02 & 1.10E-01 & 2.00E-03 & 0.00E+00 \\
cMedQA & 0.00E+00 & 2.40E-02 & 8.60E-03 & 9.40E-03 \\
\midrule
\textbf{Average} & \textbf{6.49E-03} & \textbf{2.71E-02} & \textbf{2.38E-03} & \textbf{1.56E-03} \\
\bottomrule
\end{tabular}%
    }
    \caption{The performance variation of various datasets.}
  \label{tab:variation}%
\end{table}%

\label{exp:variance}


\subsubsection{Performance Variance}
Table \ref{tab:variation} is the performance variance across the NFT, IT, MetaICL, and DrICL methods. 
We track the performance variance across all datasets with NFT(6.49E-03), IT(2.71E-02), MetaICL(2.38E-03), and DrICL(1.56E-03) as $k$ varied. 
We prove that our method demonstrates the smallest deviation, indicating greater stability in performance as $k$-shot changes.

\subsubsection{Data Noise Sensitivity}
We compare DrICL with and without local reweighting by examining how loss variance trends for each $k$-shot demonstration during training. 
The reweighting window in DrICL reduces loss variance and effectively balances the impact of noisy data by appropriately weighting demonstrations. 
This reduction in sensitivity to data noise helps maintain stable performance as the number of demonstrations increases.
For details of the noise variation please refer to Table \ref{tab:noise} in Appendix~\ref{exp:noise}.

\subsection{Ablation Studies}
\subsubsection{Hyperparameters}
Figure \ref{fig:alpha}, \ref{fig:gamma}, and \ref{fig:learning}(b) illustrate the impact of varying the hyperparameters $\alpha, \gamma$, and $S$ on the training of Llama-2-7b-chat-hf and Mistral-7B-Instruct-v0.2.
For details of the study of hyperparameters please refer to Appendix \ref{exp:parameter}.

\begin{table}[htbp]
  \centering
  \resizebox{\linewidth}{!}{
\begin{tabular}{cccccccccccc}
\toprule
\textbf{WinoWhy} & \multicolumn{1}{c}{\textbf{k=0}} & \multicolumn{1}{c}{\textbf{k=1}} & \multicolumn{1}{c}{\textbf{k=3}} & \multicolumn{1}{c}{\textbf{k=5}} & \multicolumn{1}{c}{\textbf{k=10}} & \multicolumn{1}{c}{\textbf{k=20}} & \textbf{k=30} & \textbf{k=40} & \textbf{k=50} & \multicolumn{1}{c}{\textbf{AVG}} & \multicolumn{1}{c}{\textbf{MAX}} \\
\midrule
DrICL & 0.30   & \textbf{0.51}  & \textbf{0.5}   & \textbf{0.53}  & \textbf{0.55}  & \textbf{0.57}  & 0.48  & \textbf{0.63}  & \textbf{0.52}  & \textbf{0.51}  & \textbf{0.63} \\
DrICL w/ W=1 & 0.47  & 0.46  & 0.47  & 0.51  & 0.51  & 0.49  & 0.45  & 0.58  & 0.43  & 0.49  & 0.58 \\
DrICL w/o global & 0.45  & 0.43  & 0.46  & 0.41  & 0.52  & 0.51  & 0.52  & 0.50   & 0.43  & 0.47  & 0.52 \\
DrICL w/o local & 0.43  & 0.52  & 0.44  & 0.47  & 0.51  & 0.53  & 0.33  & 0.42  & 0.33  & 0.44  & 0.52 \\
\bottomrule
\end{tabular}%
    }
    \caption{The ablation results with different settings.}
  \label{tab:window}%
\end{table}%

\subsubsection{Global and Local Contribution} Table \ref{tab:window} displays the outcomes of DrICL when applying only global strategies or local strategies exclusively to the WinoWhy dataset. The results show that refining learning objectives via a global hyperparameter to trade off the performance and the local reweighting of demonstration examples can boost the LLMs' ICL capabilities. 

\subsubsection{Analysis of Window Size} Table \ref{tab:window} shows that increasing the window size improves performance. As the sequence length grows, the number of $k$-shots also increases. Relying only on data from position $k-1$ based on previous $k-1$-shot demonstrations can cause significant variability, amplifying the impact of data fluctuations. Expanding the sampling range helps mitigate this effect. We also tested sampling from positions $0$ to $k-1$, but found the model preferentially selected certain data points, which didn't fully reflect the model's overall performance. As a result, we selected a window size of 10.

\section{Conclusions}
To enhance the ICL capacity as context lengths grow and demonstration $k$-shots rise, we introduce the DrICL algorithm to tackle the inaccurate objective and noise. 
This innovative method strategically calibrates global training goals to prioritize many-shot examples over zero-shot ones and employs local reweighting of many-shot instances using cumulative advantages as dynamic rewards, steering the model toward effective learning trajectories. 
To substantiate the effectiveness of our approach, we have curated and released the ICL-50 dataset, characterized by its diverse tasks and a broad spectrum of text lengths and quantities. 
Our method demonstrates notable enhancements in both in-domain and out-of-domain tasks. We anticipate that our research will stimulate further exploration into ICL's potential and contribute to the advancement of LLM performance.

\section*{Acknowledgments}
This work is also supported by the Public Computing Cloud, Renmin University of China and by fund for building worldclass universities (disciplines) of Renmin University of China.

\section*{Limitations}
In this work, we balance the samples in the training set, but we have not fully analyzed the algorithm's robustness across datasets of varying sizes. As a result, DrICL's performance may vary when applied to datasets of different scales, which we plan to explore in future work.
Regarding window size design, we used a uniform size for all samples. However, tasks with varying sample lengths may result in oversampling for short-text tasks and undersampling for long-text tasks. To address this, we plan to implement dynamic window sizes based on sample lengths to ensure balanced representation for both short and long samples.

\bibliography{custom}

\appendix
\clearpage

\section{Dataset}
\label{app:data}
\subsection{Overview}
ICLB dataset includes the following tasks:

$\bullet$\textbf{QA}: MMLU \cite{hendrycks2020measuring}, HellaSwag \cite{zellers2019hellaswag}, BoolQ \cite{clark2019boolq}, NarrativeQA \cite{kovcisky2018narrativeqa}, TruthfulQA \cite{lin2021truthfulqa}, OpenbookQA \cite{mihaylov2018can}, ARC \cite{clark2018think}, and QUAC \cite{choi2018quac}.

$\bullet$\textbf{Reasoning}: GSM8K \cite{cobbe2021training}, APPS \cite{hendrycks2021measuring}, MATH \cite{hendrycks2021measuring}, BABI \cite{weston2015towards}, and AR-LSAT \cite{zhong2021ar}.

$\bullet$\textbf{Summarization}: XSUM \cite{narayan2018don} and CNN/DailyMail \cite{hermann2015teaching}.

We refer to the MTEB \cite{muennighoff2022mteb} and C-MTEB \cite{xiao2023c} that contains the description of the following dataset.

$\bullet$\textbf{Clustering}: ArxivClusteringS2S, ArxivClusteringP2P,
BiorxivClusteringS2S, BiorxivClusteringP2P,
MedrxivClusteringP2P, RedditClustering, RedditClusteringP2P, StackExchangeClustering, StackExchangeClusteringP2P, CLSClusteringS2S, CLSClusteringP2P, ThuNewsClusteringS2S, BlurbsClusteringS2S, BlurbsClusteringP2P, TenkgnadClusteringS2S, TenkgnadClusteringP2P.

$\bullet$\textbf{Classification}: AmazonPolarity, AmazonReviews, Emotion, ToxicConversations, TweetSentimentExtraction, JDReview, MultilingualSentiment, OnlineShopping, Waimai and WinoWhy \cite{zhang2020winowhy}.

$\bullet$\textbf{Retrieval}: cMedQA, TREC-COVID, DuReaderRetrieval, EcomRetrieval, MMarco, MedicalRetrieval, T2R, and VideoRetrieval.

$\bullet$\textbf{Reranking}: cMedQA and AskUbuntuDupQuestions.

\subsection{Data Analysis}
\begin{figure*}[t]
    \centering
    \includegraphics[width=0.9\textwidth]{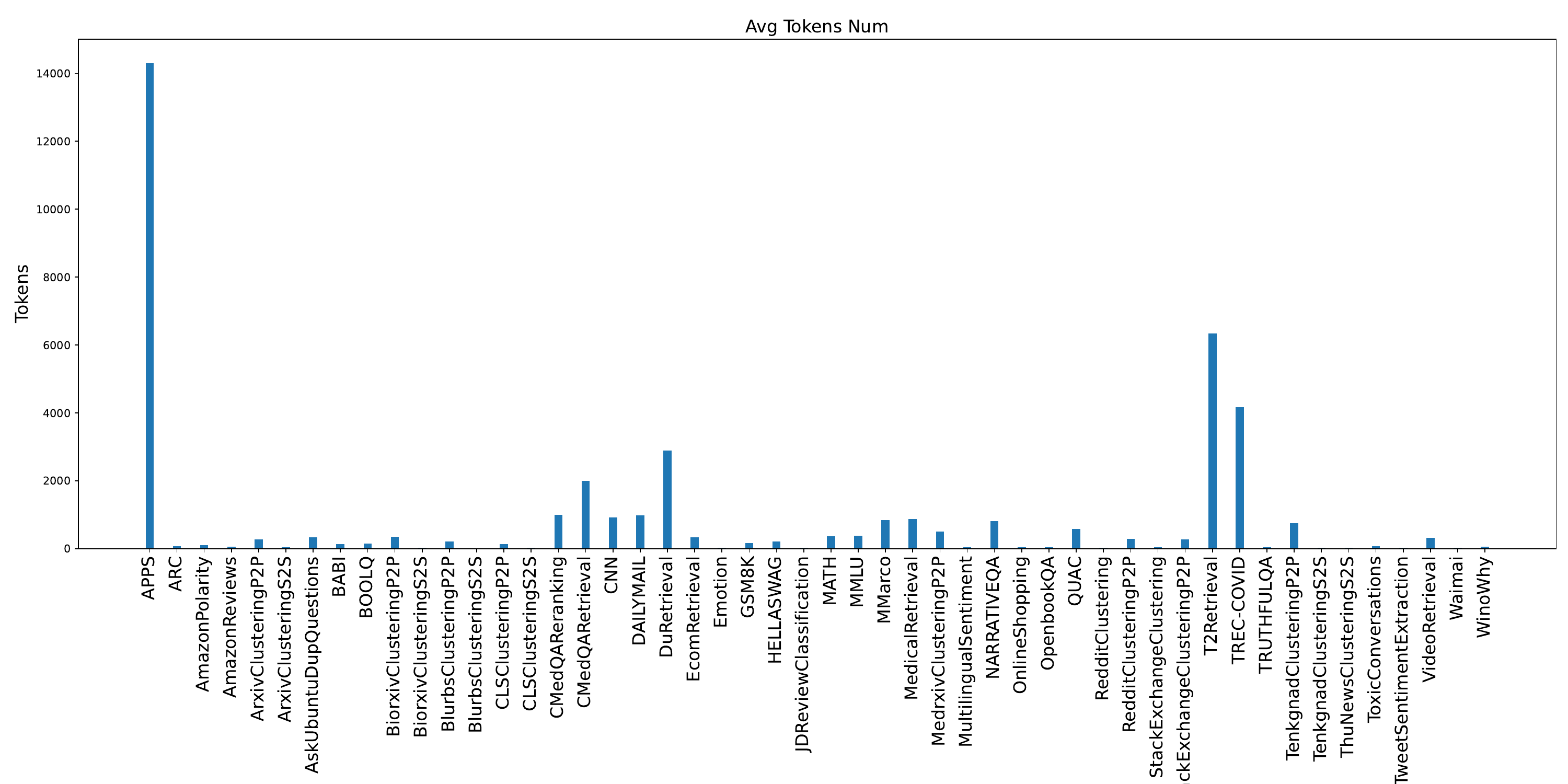} 
    \caption{The token distributions of each task dataset.}
    \label{fig:token}
    \vspace{-0.5cm}
\end{figure*}

\begin{figure*}[t]
    \centering
    \includegraphics[width=0.9\textwidth]{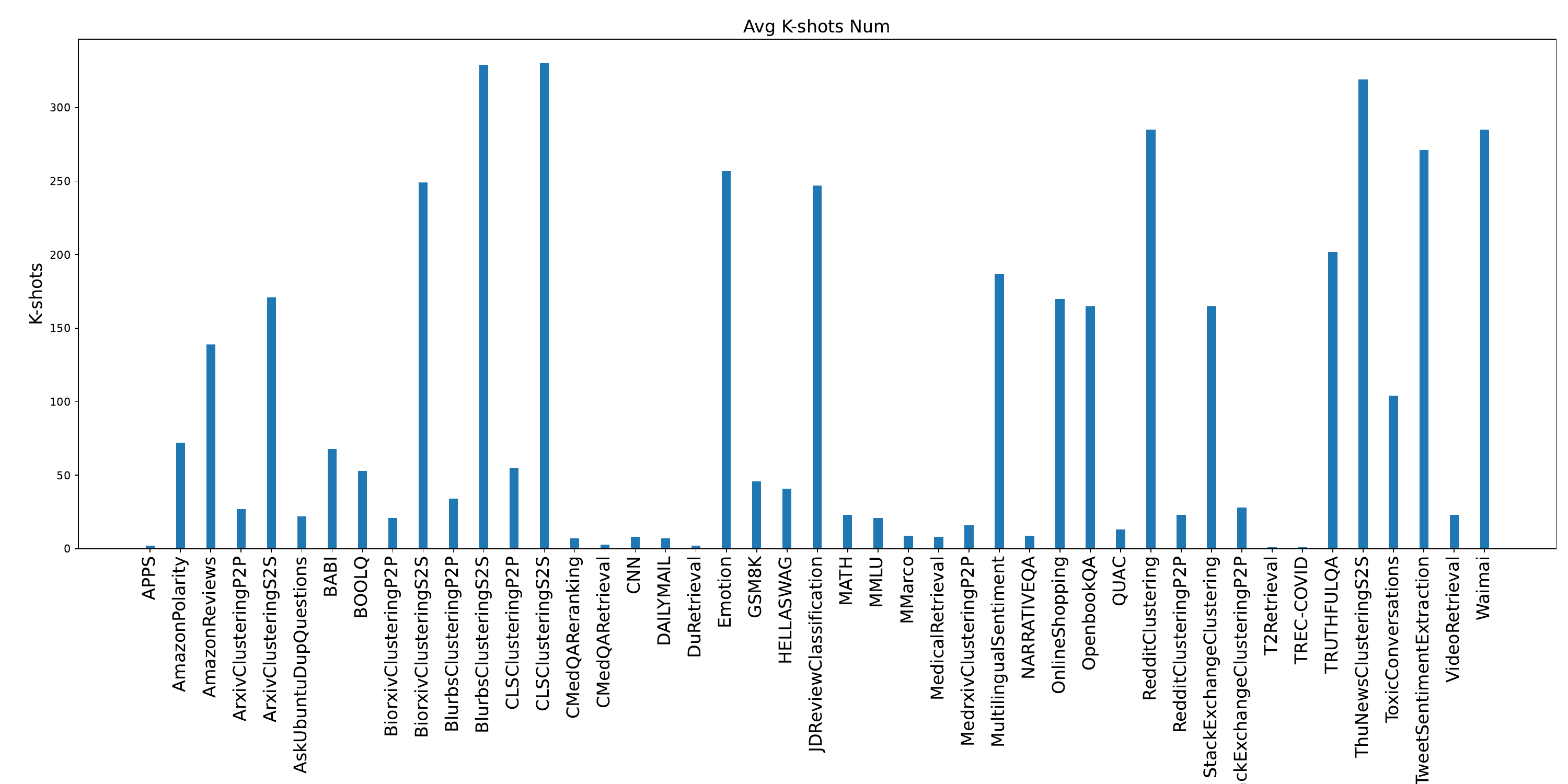} 
    \caption{The $k$-shots distributions of each task dataset.}
    \label{fig:shots}
\end{figure*}

The statistics of data volume for each task can be referred to in Table \ref{tab:statistics}, where the data volume for 50 tasks ranges from several hundred to hundreds of thousands of entries, providing ample data for the model's training and inference. 
The distribution of the number of tokens for tasks is between 10 and 14,000, as shown in Figure \ref{fig:token}. 
When the many-shot fine-tuning sequence length is fixed, the $k$-shot number varies significantly. 
Figure \ref{fig:shots} illustrates the $k$-shot distribution with a fixed training sequence length of 8,000, ranging between 0 and 350.
When the training sequence length is increased to 32,000, the range of $k$-shot variation will exceed 1,000. 
This provides a solid data foundation for the performance study of ICL in many-shot scenarios.

\begin{table}[htbp]
\small
  \centering
  \resizebox{\linewidth}{!}{
\begin{tabular}{cccc}
\toprule
\textbf{Task Type} & \textbf{Task Name} & \textbf{Train} & \textbf{Test} \\
\midrule
\multirow{8}[2]{*}{QA} & MMLU  & 99834 & 13985 \\
      & HellaSwag & 39905 & 10042 \\
      & BoolQ & 9427  & 100 \\
      & NarrativeQA & 36208 & 0 \\
      & TruthfulQA & 22434 & 0 \\
      & OpenbookQA & 4957  & 500 \\
      & QUAC  & 83568 & 7354 \\
      & ARC   & 1096  & 106 \\
\midrule
\multirow{5}[2]{*}{Reasoning} & APPS  & 5000  & 0 \\
      & MATH  & 7500  & 5000 \\
      & BABI  & 200000 & 20000 \\
      & GSM8K & 7473  & 1319 \\
      & AR-LSAT & 1630  & 230 \\
\midrule
\multirow{3}[2]{*}{Summarization} & CNN   & 83321 & 9258 \\
      & DailyMail & 197555 & 21951 \\
      & XSUM  & 204045 & 11334 \\
\midrule
\multirow{16}[2]{*}{Clustering} & CLSClusteringP2P & 81499 & 8501 \\
      & CLSClusteringS2S & 83359 & 6641 \\
      & ThuNewsClusteringS2S & 83816 & 6184 \\
      & ArxivClusteringP2P & 135171 & 14829 \\
      & ArxivClusteringS2S & 133190 & 16810 \\
      & BiorxivClusteringP2P & 58185 & 6815 \\
      & BiorxivClusteringS2S & 57893 & 7107 \\
      & BlurbsClusteringP2P & 135018 & 14982 \\
      & BlurbsClusteringS2S & 132972 & 17028 \\
      & MedrxivClusteringP2P & 29337 & 3163 \\
      & RedditClustering & 134749 & 15251 \\
      & RedditClusteringP2P & 134391 & 15609 \\
      & StackExchangeClustering & 132996 & 17004 \\
      & StackExchangeClusteringP2P & 58584 & 6416 \\
      & TenkgnadClusteringP2P & 38953 & 4110 \\
      & TenkgnadClusteringS2S & 39293 & 3770 \\
\midrule
\multirow{10}[2]{*}{Classification} & JDReview & 3468  & 261 \\
      & MultilingualSentiment & 90761 & 9239 \\
      & OnlineShopping & 7675  & 325 \\
      & AmazonPolarity & 89979 & 10021 \\
      & AmazonReviews & 90145 & 9855 \\
      & Emotion & 12124 & 3876 \\
      & ToxicConversations & 45823 & 4177 \\
      & TweetSentimentExtraction & 24189 & 3292 \\
      & Waimai & 7697  & 303 \\
      & WinoWhy & 1160  & 443 \\
\midrule
\multirow{8}[2]{*}{Retrieval} & cMedQARetrieval & 5898  & 804 \\
      & TREC-COVID & 803   & 79 \\
      & DuReaderRetrieval & 7996  & 865 \\
      & EcomRetrieval & 757   & 139 \\
      & MMarco & 5944  & 741 \\
      & MedicalRetrieval & 829   & 78 \\
      & T2R   & 8958  & 1042 \\
      & VideoRetrieval & 859   & 28 \\
\midrule
\multirow{2}[2]{*}{Reranking} & cMedQAReranking & 806   & 99 \\
      & AskUbuntuDupQuestions & 295   & 45 \\
\bottomrule
    \end{tabular}}%
    \caption{The statistics of each task dataset.}
  \label{tab:statistics}%
\end{table}%

\subsection{Data Deploy}
For each task, we leverage GPT-3.5-Turbo to generate instructions.
We segment our datasets into meta-train and meta-test sets, holding back data from one task per category for evaluating our method's ability to generalize to new data. 
For overall evaluation, we possess both in-domain and out-of-domain test sets in comparison to the meta-train data. 
For Classification, meta-train domains include online shopping, multilingual, sentiment analysis, and toxicity detection, while test domains extend to dining and common sense. 
For Reasoning, math-related domains are included in both meta-train and test sets.
When assembling the training set, we ensure an equitable distribution of data among tasks, keeping the difference in data volume between any two tasks to no more than ten times, thereby enhancing model performance.
We generate demonstrations with varying $k$-shots using the training set. 
For each task, we infer 100 randomly selected test set entries per $k$-shot, assessing the model's performance with different $k$-shot ranges from 0 to 350.

\section{Experiment Details}
\label{exp:eval}
\subsection{Evaluation}
We evaluated our method on 12 datasets, each with 1,600 samples, totaling 19,200 test samples, and sampling rates ranging from 2\% to 100\%. For each dataset, we collected 16 types of demonstrations with various $k$-shot values, including 0, 1, 3, 5, 10, 20, 30, 40, 50, 60, 70, 80, 90, 100, 200, and 300. For reference, Table \ref{tab:total_test} presents the results from Table \ref{tab:all} for the total test set. The findings show that performance trends remain consistent across different sampling instances from each dataset.

\begin{table*}[h]
  \small
  \centering
  \resizebox{15cm}{!}{
\begin{tabular}{ccccccccccccccc}
\toprule
\multicolumn{1}{c}{\textbf{Dataset}} & \textbf{Models} & \multicolumn{1}{c}{\textbf{k=0}} & \multicolumn{1}{c}{\textbf{k=1}} & \multicolumn{1}{c}{\textbf{k=3}} & \multicolumn{1}{c}{\textbf{k=5}} & \multicolumn{1}{c}{\textbf{k=10}} & \multicolumn{1}{c}{\textbf{k=20}} & \multicolumn{1}{c}{\textbf{k=30}} & \multicolumn{1}{c}{\textbf{k=40}} & \multicolumn{1}{c}{\textbf{k=50}} & \multicolumn{1}{c}{\textbf{k=60}} & \multicolumn{1}{c}{\textbf{k=70}} & \multicolumn{1}{c}{\textbf{AVG}} & \multicolumn{1}{c}{\textbf{MAX}} \\
\midrule
\multicolumn{1}{c}{\multirow{4}[2]{*}{OpenbookQA$\stackrel{\text{id}}{}$}} & NFT   & 0.29  & 0.25  & 0.24  & 0.23  & 0.27  & 0.26  & 0.19  & 0.16  & 0.2   & 0.18  & 0.18  & 0.22  & 0.29 \\
      & IT    & 0.71  & 0.51  & 0.58  & 0.6   & 0.6   & 0.64  & 0.62  & 0.58  & 0.58  & 0.53  & 0.52  & 0.59  & 0.71 \\
      & MetaICL & 0.63  & 0.65  & 0.68  & 0.69  & 0.73  & 0.75  & 0.76  & 0.76  & 0.76  & 0.76  & 0.74  & 0.72  & 0.76 \\
      & \textbf{DrICL} & \textbf{0.74} & \textbf{0.74} & \textbf{0.77} & \textbf{0.76} & \textbf{0.77} & \textbf{0.79} & \textbf{0.78} & \textbf{0.79} & \textbf{0.79} & \textbf{0.78} & \textbf{0.77} & \textbf{0.77} & \textbf{0.79} \\
\midrule
\multicolumn{1}{c}{\multirow{4}[2]{*}{CLSClusteringS2S$\stackrel{\text{id}}{}$}} & NFT   & 0.17  & 0.03  & 0.01  & 0     & 0     & 0     & 0.02  & 0.03  & 0.01  & 0.02  & 0.12  & 0.04  & 0.17 \\
      & IT    & 0.86  & 0.81  & 0.74  & 0.71  & 0.66  & 0.53  & 0.52  & 0.49  & 0.53  & 0.52  & 0.51  & 0.63  & 0.86 \\
      & MetaICL & 0.82  & 0.85  & 0.84  & 0.86  & 0.85  & 0.85  & 0.83  & 0.85  & 0.85  & 0.84  & 0.83  & 0.84  & 0.86 \\
      & \textbf{DrICL} & 0.85  & \textbf{0.86} & \textbf{0.86} & 0.84  & \textbf{0.87} & \textbf{0.86} & \textbf{0.86} & \textbf{0.87} & \textbf{0.85} & \textbf{0.86} & \textbf{0.88} & \textbf{0.86} & \textbf{0.88} \\
\midrule
\multicolumn{1}{c}{\multirow{4}[2]{*}{ArxivClusteringS2S$\stackrel{\text{ood}}{}$}} & NFT   & 0.02  & 0.03  & 0.06  & 0.06  & 0.05  & 0.05  & 0.03  & 0.03  & 0.04  & 0.06  & 0.02  & 0.04  & 0.06 \\
      & IT    & 0.36  & 0.32  & 0.25  & 0.29  & 0.25  & 0.22  & 0.22  & 0.23  & 0.21  & 0.19  & 0.16  & 0.25  & 0.36 \\
      & MetaICL & 0.35  & 0.39  & 0.35  & 0.35  & 0.34  & 0.37  & 0.37  & 0.39  & 0.38  & 0.36  & 0.33  & 0.36  & 0.39 \\
      & \textbf{DrICL} & 0.33  & 0.34  & \textbf{0.37} & \textbf{0.36} & \textbf{0.36} & \textbf{0.39} & \textbf{0.4} & 0.38  & 0.36  & \textbf{0.42} & \textbf{0.34} & \textbf{0.37} & \textbf{0.42} \\
\bottomrule
\end{tabular}%
    }
    \caption{The evaluation on the whole test set of OpenbookQA, CLSClusteringS2S, and ArxivClusteringS2S. 
    \textbf{Bold} indicates that our model performs the best.
}
  \label{tab:total_test}%
\end{table*}%

\label{exp:parameter}
\subsection{Hyperparameters Study}
\begin{figure}[htb]
    \centering
\includegraphics[width=1.0\linewidth]{
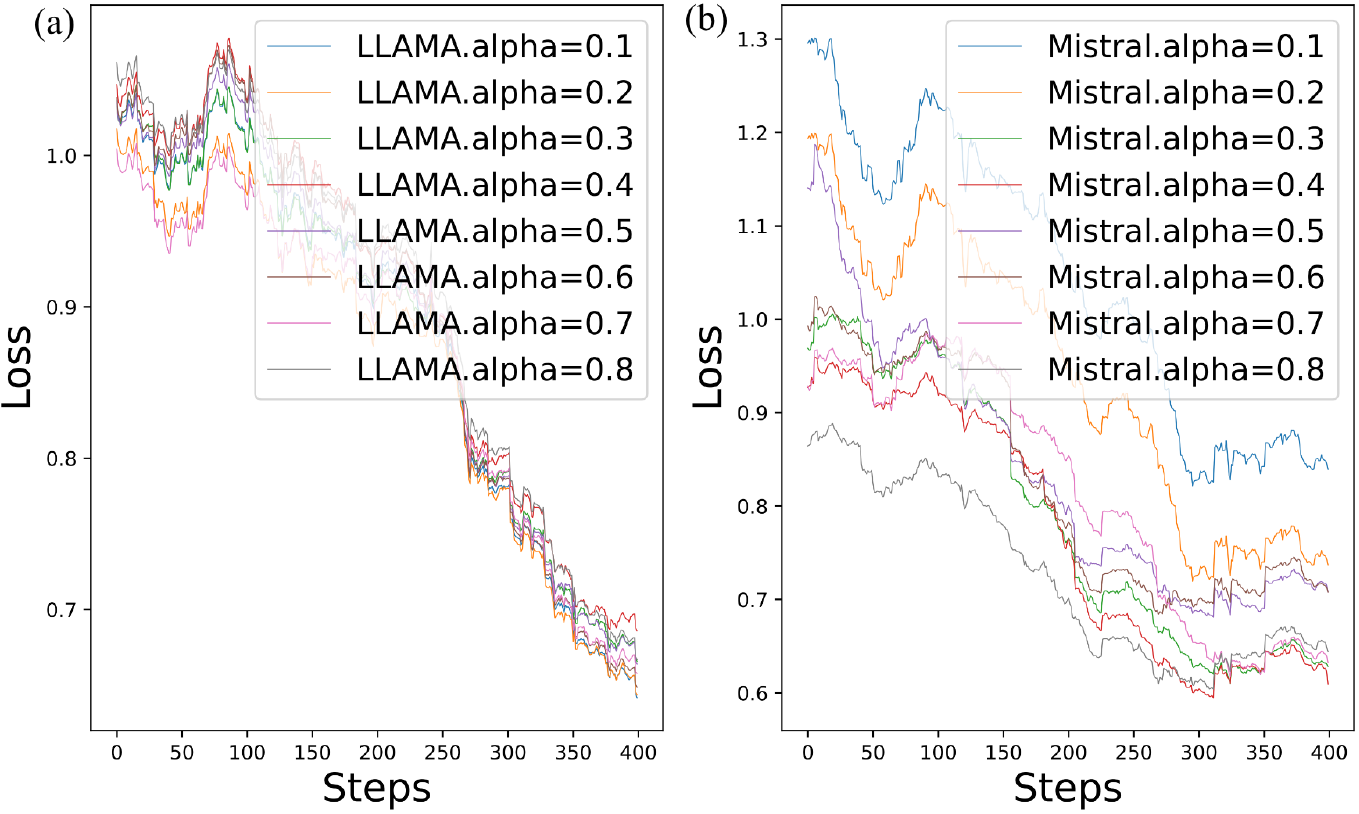} 
    \caption{(a) The optimal performance is when the parameter $\alpha$ is set to 0.2 on Llama-2-7b-chat-hf. 
    (b) The optimal performance on Mistral-7B-Instruct-v0.2 is achieved with $\alpha$ set to 0.4.}
    \label{fig:alpha}
\end{figure}

$\alpha$ plays a crucial role in determining the model's performance, as illustrated in Figure \ref{fig:alpha}, the variance in performance between zero-shot and many-shot scenarios is model-dependent. 

\begin{figure}[htb]
    \centering
\includegraphics[width=1.0\linewidth]{
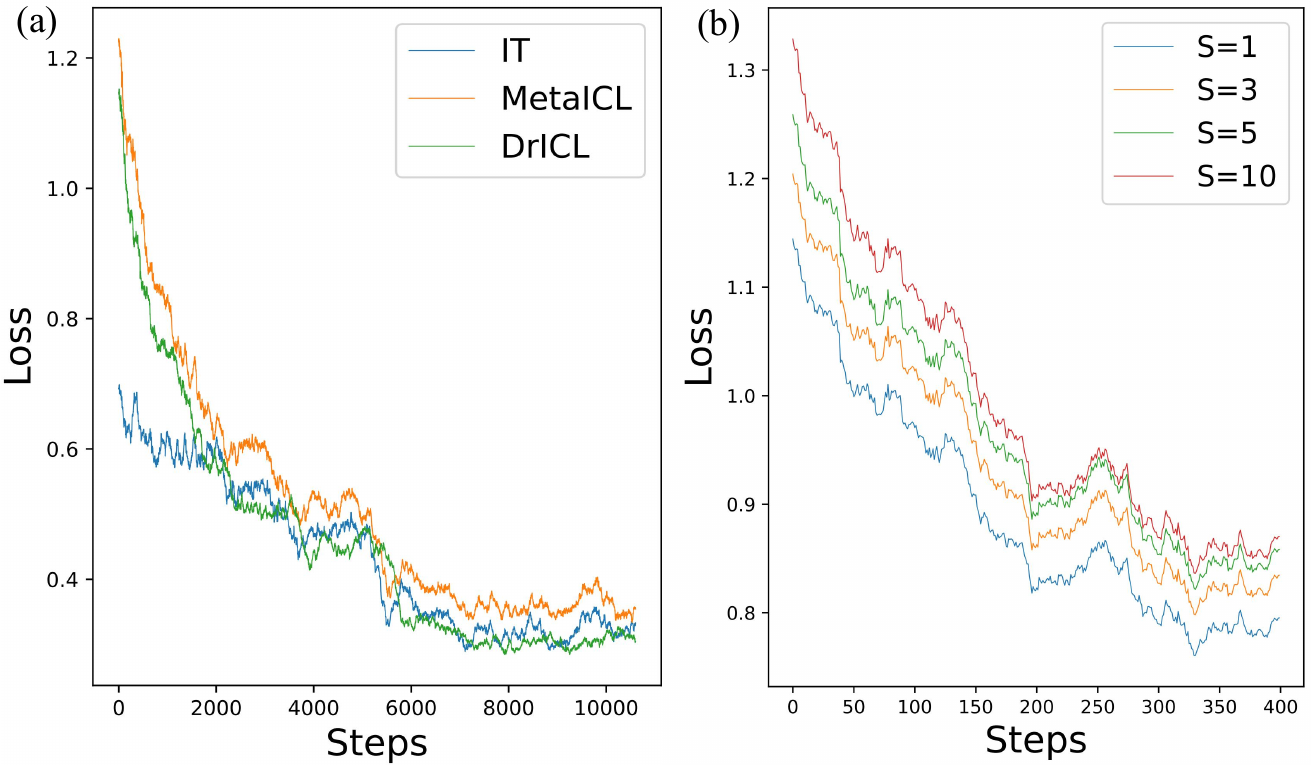} 
    \caption{(a) The many-shot training loss of DrICL converges to a lower level compared to IT and MetaICL. (b) The optimal performance is when the parameter $S$ is set to 1 on Llama-2-7b-chat-hf.}
    \label{fig:learning}
\end{figure}

\begin{figure}[htb]
    \centering
\includegraphics[width=1.0\linewidth]{
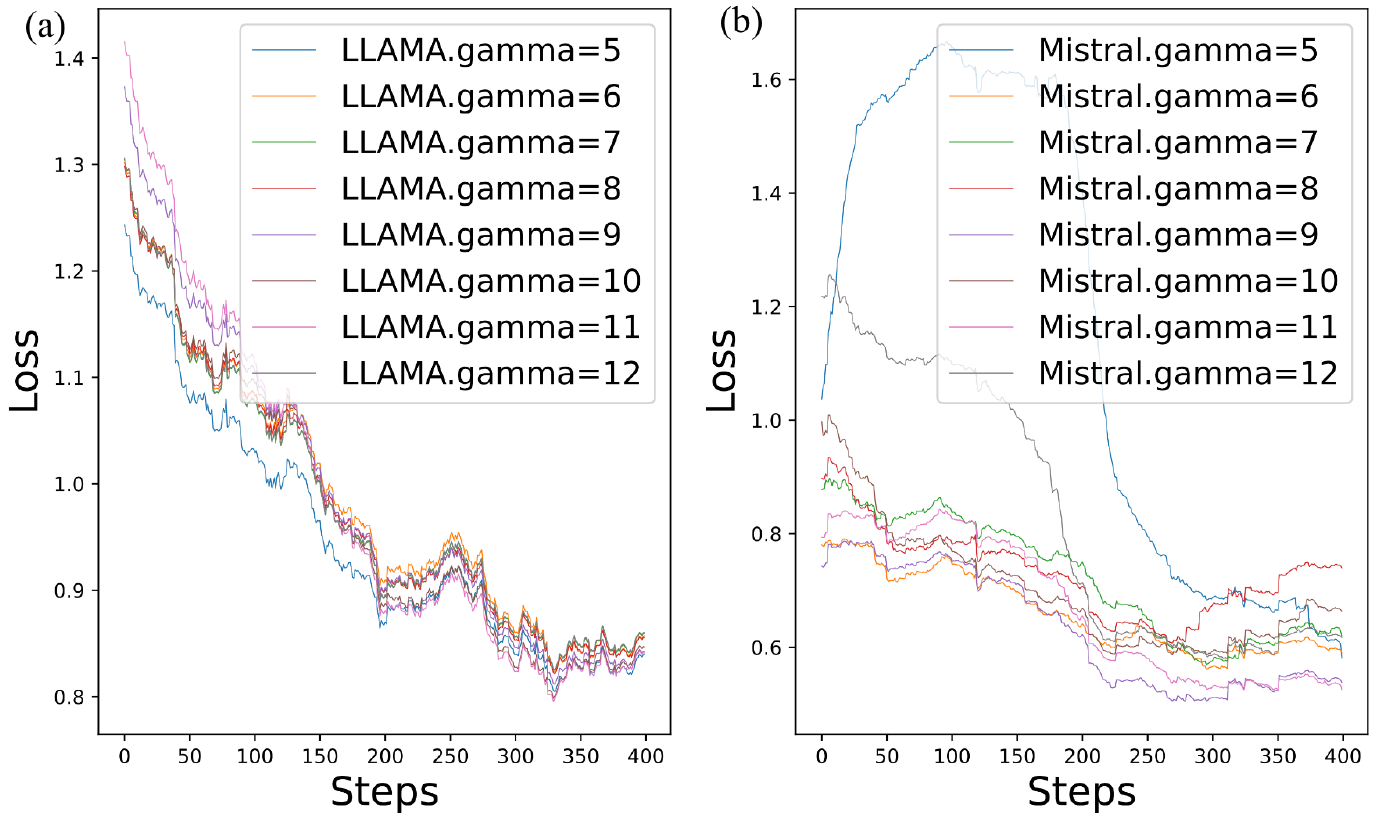} 
    \caption{(a) and (b) show the optimal $\gamma$ settings for Llama-2-7b-chat-hf and Mistral-7B-Instruct-v0.2, with both models achieving the best performance at $\gamma = 11$.}
    \label{fig:gamma}
\end{figure}

$\gamma$ adjusts the sensitivity of the rewards and makes the training process stable. Figure \ref{fig:gamma} illustrates that the best setting of $\gamma$ is 11.


$S$ represents the loss computed from three randomly sampled positions within the sampling window to calculate the reward. A high value of $S$ leads to non-representative sampling. From our experiments in Figure \ref{fig:learning}(b), we found that the best training results were achieved with $S=1$.

\label{exp:noise}
\subsection{Data Noise Sensitivity}
Table \ref{tab:noise} illustrates the loss variance at different training stages with and without local reweighting.

\begin{table}[htbp]
  \centering
  \resizebox{\linewidth}{!}{
\begin{tabular}{cccccc}
\toprule
\multirow{2}[2]{*}{\textbf{Methods}} & \multicolumn{5}{c}{\textbf{Variance Across Training Progress}} \\
      & \textbf{20\%} & \textbf{40\%} & \textbf{60\%} & \textbf{80\%} & \textbf{100\%} \\
\midrule
\textbf{DrICL w/o Local} & 4.90   & 1.40   & 0.93  & 0.59  & 1.80 \\
\midrule
\textbf{DrICL} & 6.60   & 1.85  & 0.55  & 0.35  & 0.32 \\
\bottomrule
\end{tabular}%
    }
  \caption{The loss variance during the whole training process.}
  \label{tab:noise}%
\end{table}%


\end{document}